\documentclass[conference]{IEEEtran}
\IEEEoverridecommandlockouts
\usepackage{cite}
\usepackage{amsmath,amssymb,amsfonts}
\usepackage{algorithmic}
\usepackage{graphicx}
\usepackage{makecell}
\usepackage{booktabs}
\graphicspath{ {./Figures/} }
\usepackage{textcomp}
\usepackage{xcolor}
\usepackage{comment}
\usepackage{authblk}
\usepackage{multirow}
\def\BibTeX{{\rm B\kern-.05em{\sc i\kern-.025em b}\kern-.08em
    T\kern-.1667em\lower.7ex\hbox{E}\kern-.125emX}}

\begin{document}
\title{Multimodal Deep Learning for Phyllodes Tumor Classification from Ultrasound and Clinical Data\\
}
\author[1]{Farhan Fuad Abir}
\author[3]{Abigail Elliott Daly}
\author[3]{Kyle Anderman}
\author[3]{Tolga Ozmen}
\author[2]{Laura J. Brattain}

\affil[1]{Department of Electrical and Computer Engineering, University of Central Florida, Orlando, FL, USA}
\affil[2]{Department of Medicine, University of Central Florida College of Medicine, Orlando, FL, USA}
\affil[3]{Massachusetts General Hospital, Department of Surgery, Section of Breast Surgery, MA, USA}

\maketitle

\begin{abstract}
    Phyllodes tumors (PTs) are rare fibroepithelial breast lesions that can be malignant but are difficult to classify preoperatively due to their radiological similarity to benign fibroadenomas. This often leads to unnecessary surgical excisions. To address this, we propose a multimodal deep learning framework that integrates breast ultrasound (BUS) images with structured clinical data to improve diagnostic accuracy. We developed a dual-branch neural network that extracts and fuses features from ultrasound images and patient metadata from 81 subjects with confirmed PTs. Class-aware sampling and subject-stratified 5-fold cross-validation were applied to mitigate class imbalance and data leakage. The results show that our proposed multimodal method outperforms unimodal baselines in classifying benign versus borderline/malignant PTs. Among six deep learning-based image encoders, ConvNeXt and ResNet18 achieved the best performance in the multimodal setting, with AUC-ROC scores of 0.9427 and 0.9349, and F1-scores of 0.6720 and 0.7294, respectively. This study demonstrates the potential of multimodal AI to serve as a non-invasive diagnostic tool, reducing unnecessary surgical excisions and improving clinical decision-making in breast cancer care.
\end{abstract}


\section{Introduction}

Breast fibroepithelial lesions are a diverse group of biphasic tumors consisting of fibroadenomas (FAs) and phyllodes tumors (PTs). While FAs are common benign tumors with a low risk of recurrence or metastasis, PTs are rare lesions, accounting for less than 1\% of breast tumors, that exhibit heterogeneous characteristics \cite{soleimani2024escalating}. However, their clinical significance lies in the uncertain progression and the diagnostic challenges \cite{lissidini2022malignant}. Moreover, the World Health Organization (WHO) classifies phyllodes tumors into benign, borderline, and malignant subtypes with reported local recurrence rates of 7.1\%, 16.7\%, and 25.1\%, respectively \cite{yan2025hierarchical}. Therefore, accurate preoperative grading of PTs into benign versus borderline/malignant is crucial for appropriate treatment planning.

The current standard diagnostic pathway often involves a core needle biopsy (CNB), followed by surgical excision if the pathology is ambiguous. However, only up to 57\% of CNB-diagnosed fibroepithelial lesions are later upgraded to phyllodes tumors after excision, leading to unnecessary surgeries with potential complications, scarring, and increased healthcare costs. In the United States alone, excisional biopsies for suspected phyllodes tumors cost over \$2 million annually \cite{mousa2022excision, soleimani2024escalating}. Artificial Intelligence (AI) could potentially mitigate this challenge by improving the differentiation of malignant PTs from the benign ones.

Early efforts focused on ultrasound (US) imaging alone. Yan et al. developed a deep learning (DL) model differentiating PTs from fibroadenomas (FAs) with 87.3\% accuracy using multicenter ultrasound data \cite{yan2025hierarchical}. Similarly, the Xception architecture achieved an area under the curve (AUC) of 0.87 for PT-FA discrimination \cite{yan2024deep}. However, these models lacked clinical context, limiting their diagnostic utility in ambiguous cases \cite{pfob2022importance}.

Multimodal AI approaches often integrate imaging with clinical data, and they show significant promise in broader breast cancer diagnostics. Ben et al. incorporated patient metadata with mammography, a tissue characterization technique, which improved molecular subtyping AUC by up to 27.6\% over image-only models \cite{ben2025multimodal}. Similarly, fusing B-mode ultrasound with Nakagami parametric imaging increased AUC by 10.75\% \cite{muhtadi2025breast}, and combining mammography and ultrasound boosted specificity to 96.41\% \cite{chen2025deep}. Xu et al. integrated morphological and texture features that reduced false positives in BI-RADS 4 lesions by 15\% \cite{xu2024classification}. Despite the progress, multimodal strategies have rarely been applied to phyllodes tumors, where models continue to rely on ultrasound or histopathological features alone. This is a critical gap, as clinical data, such as age, tumor size, and menopausal status, are central to the overall evaluation and could significantly enhance diagnostic performance in AI models.

In this work, we propose a multimodal deep learning framework integrating US images and structured clinical data for PT classification. Our dual-branch architecture fuses image and tabular features, while class-aware sampling and subject-stratified 5-fold cross-validation address class imbalance and data leakage. We also evaluated the method on multiple deep learning-based image encoders, all of which outperformed unimodal baselines, highlighting its potential as a non-invasive decision support tool in clinical workflows.

\section{Methods} 

\subsection{Dataset}

The phyllodes tumor dataset was retrospectively collected and anonymized at the Massachusetts General Hospital (MGH) and was approved by the Institutional Review Board. The deidentified dataset comprises clinical data from 106 subjects, breast ultrasound (BUS) images from 71, and BUS videos from 10 subjects in DICOM format. Not all subjects had complete clinical and imaging data. After data matching, a subset of 81 subjects (an average age of 38.01 $\pm$ 13.37 years) with both modalities was selected for this study. In terms of phyllodes tumor class, there were 65 benign, 10 borderline, and 6 malignant subjects. For this study, we combined the borderline and malignant cases to formulate a binary classification task. 

\subsection{Preprocessing}
The BUS images and videos, originally in DICOM format, were converted to PNG format. Duplicate images were removed, and only frames containing tumors were retained. Since most BUS images included text along the image borders (e.g., probe position, frequency, and other metadata), we cropped the borders to remove these artifacts. Additionally, the extracted images were in RGB, and we converted them to grayscale to ensure a consistent color profile across the dataset. The resulting image dataset consisted of 1,213 benign and 425 malignant images, resulting in a class ratio of approximately 3:1.

The clinical dataset was then preprocessed based on feature availability, retaining three numeric (age at diagnosis, BMI, and tumor size) and three categorical features (race, menopausal status, and tumor echogenicity). Categorical variables were processed using one-hot encoding, yielding a total of 10 clinical features.

\subsection{Class-aware sampling}

In highly imbalanced datasets, conventional random sampling often results in mini-batches dominated by the majority class, which can lead to biased model updates and poor generalization to minority classes. 
Class-aware sampling addresses this issue by decoupling the sampling process into two stages. First, a class is selected uniformly at random, ensuring equal sampling probability for each class regardless of its prevalence in the dataset. Second, a sample is drawn uniformly from within the selected class. By repeating this process, the constructed mini-batches contain a more equitable distribution of classes over time \cite{shen2016relay}. This approach has effectively improved the training dynamics of deep learning models on long-tailed and imbalanced datasets \cite{varaganti2025t2id}. 

\subsection{Multimodal Deep Learning Framework}

\begin{figure}[ht]
    \centering
    \includegraphics[width=0.9\linewidth]{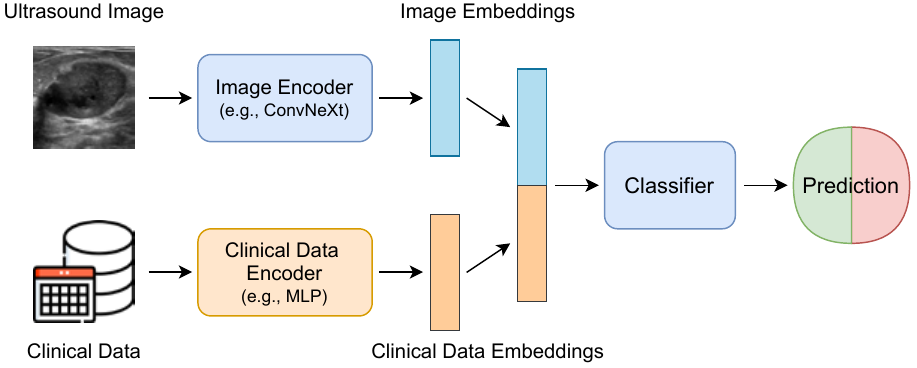}
    \caption{Multimodal AI model architecture with feature-level fusion. The image branch consists of an image encoder (e.g., VGG16, ConvNeXt), while the clinical data branch uses an MLP. The resulting embeddings are concatenated and passed to a deep learning classifier (fully-connected layer) for the final prediction.}
    \label{fig:multimodal_diag}
\end{figure}

Our multimodal architecture consists of two parallel branches as illustrated in Figure \ref{fig:multimodal_diag}. 

The imaging branch is a convolutional neural network (CNN)-based image encoder to extract one-dimensional feature embeddings from grayscale ultrasound images. Transfer learning was used during training to finetune the ImageNet-pretrained image encoders. In parallel, the clinical data (10-dimensional input) was processed by a two-layer multilayer perceptron (MLP) with 10$\rightarrow$16$\rightarrow$8 units and ReLU activations to generate a compact embedding of the tabular features. The two embeddings were then concatenated at the feature level. Lastly, the fused vector was passed through a final linear classifier with a softmax activation function to predict the class label.


This design enables the model to jointly reason over visual and tabular inputs, enhancing robustness and interpretability.

\section{Results}

\subsection{Experimental Setup}

We performed subject-stratified 5-fold cross-validation to prevent data leakage across training and validation sets. Moreover, class-aware sampling was applied during training to address class imbalance. Ultrasound images were resized to 224$\times$224 pixels, and the intensity values were normalized to the [0, 1] range. During training, we applied random horizontal flipping and small rotation ($\pm10^\circ$). Models were trained for 30 epochs using stochastic gradient descent (SGD) with a learning rate of 0.001, momentum of 0.9, and a batch size of 4. Each model was evaluated using average accuracy, F1-score, and its 95\% confidence interval (CI) across folds. All experiments were conducted using an NVIDIA RTX A4000 GPU (16GB VRAM) and a system with 32GB RAM.

\subsection{Evaluation of multimodal image encoders}

We evaluated six ImageNet-pretrained image encoders: VGG16 \cite{simonyan2014very}, ResNet18, ResNet50 \cite{he2016deep}, EfficientNet-B7 \cite{tan2019efficientnet}, CCT \cite{hassani2021escaping}, and ConvNeXt \cite{liu2022convnet} within the multimodal architecture. Table~\ref{tab:encoders} summarizes the average accuracy, F1-score, and AUC-ROC over five subject-stratified cross-validation folds. 

\begin{table}[!ht]
\caption{Evaluation of the multimodal framework with different image encoders (sorted by AUC-ROC)}
\label{tab:encoders}
\begin{tabular}{@{}ccccc@{}}
\toprule
\begin{tabular}[c]{@{}c@{}}Image\\ Encoder\end{tabular} &
\begin{tabular}[c]{@{}c@{}}Mean\\ Accuracy\end{tabular} &
\begin{tabular}[c]{@{}c@{}}Mean\\ F1-Score\end{tabular} &
\begin{tabular}[c]{@{}c@{}}Mean\\ AUC\end{tabular} &
\begin{tabular}[c]{@{}c@{}}95\% CI\\ (AUC)\end{tabular} \\ \midrule
ConvNeXt        & 0.9098 & 0.6720 & \textbf{0.9427} & \textbf{[0.8842, 1.0000]} \\
CCT             & 0.9018 & 0.6683 & 0.9406 & [0.8633, 1.0000] \\
ResNet18        & \textbf{0.9179} & \textbf{0.7294} & 0.9349 & [0.8775, 0.9923] \\
ResNet50        & 0.9100 & 0.7065 & 0.9168 & [0.8253, 1.0000] \\
EfficientNet-B7 & 0.8252 & 0.5473 & 0.8963 & [0.8203, 0.9722] \\
VGG16           & 0.8885 & 0.6389 & 0.8886 & [0.7376, 1.0000] \\
\bottomrule
\end{tabular}
\end{table}

Among all models, \textbf{ConvNeXt} achieved the highest AUC-ROC of 0.9427, closely followed by CCT (0.9406) and ResNet18 (0.9349). In addition, ConvNeXt achieved a sensitivity of 0.71, a specificity of 0.95, a PPV of 0.87, and an NPV of 0.81. \textbf{ResNet18} obtained the best average accuracy of 0.9179 and an F1-score of 0.7294. In contrast, \textbf{EfficientNet-B7} and \textbf{VGG16} exhibited lower performance, particularly in F1-score (0.5473 and 0.6389, respectively), likely reflecting their sensitivity to class imbalance or limited generalization on small datasets.

\subsection{Attribution analysis}

\begin{figure}[!ht]
    \centering
    \includegraphics[width=1\linewidth]{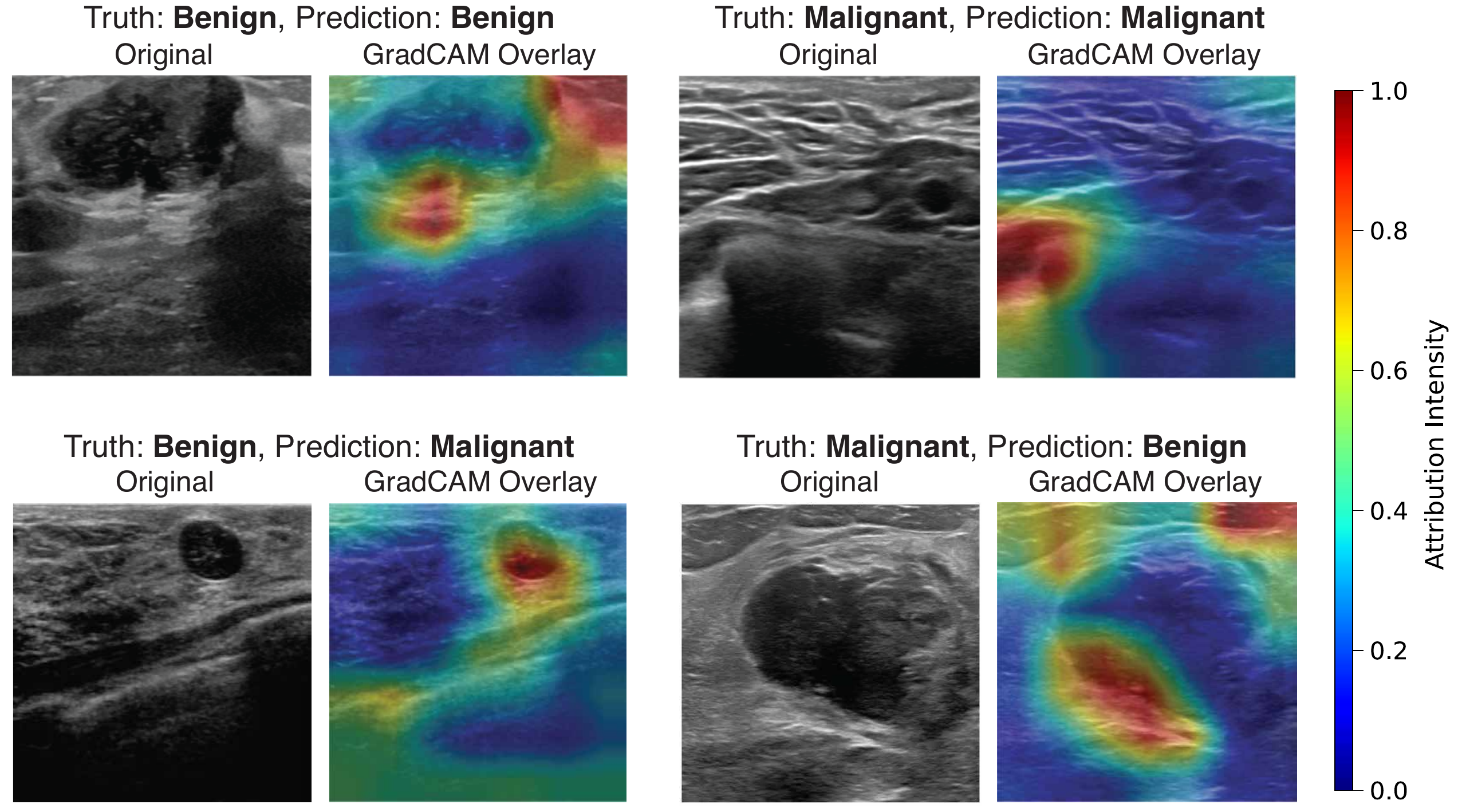}
    \caption{Score-CAM attributes of the ConvNeXt image encoder benign and malignant cases. Warmer colors indicate regions contributing more strongly to the predicted class. Correctly predicted cases show focused attention on relevant tumor regions, whereas misclassified cases often exhibit diffuse or misplaced activations, suggesting visual ambiguity or model uncertainty.}
    \label{fig:CAM}
\end{figure}

To improve explainability, we used Score-CAM to visualize salient regions in ultrasound images, highlighting the model’s decision process. Score-CAM generates class-specific attribution maps by weighting activation maps with their corresponding class scores \cite{wang2020score}. Figure~\ref{fig:CAM} shows examples of true positive, true negative, and misclassified cases using ConvNeXt as the image encoder.

In the correctly predicted malignant cases, the image encoder focused on tumor regions and the surroundings with heterogeneous echotexture, while the misclassified samples showed dispersed or ambiguous activation patterns. These insights suggest that visual cues alone are sometimes insufficient, which further justifies the use of multimodal reasoning.

\subsection{Modality Ablation}

\begin{figure}[!ht]
    \centering
    \includegraphics[width=0.7\linewidth]{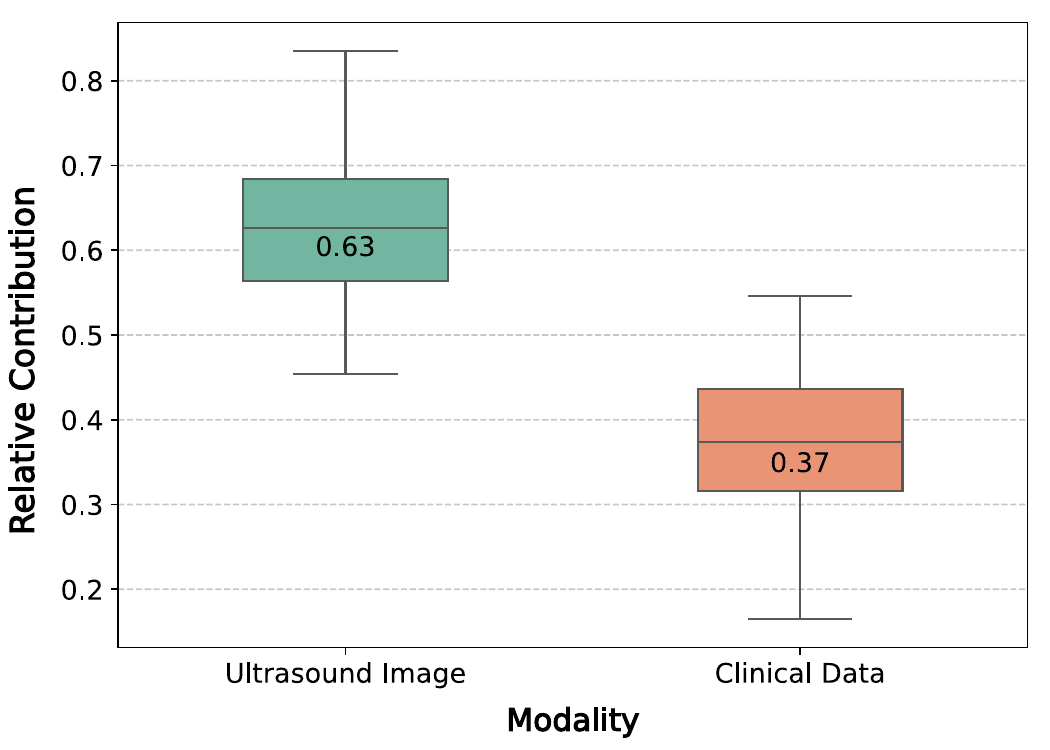}
    \caption{Modality ablation analysis showing relative contributions of ultrasound images and clinical data in the multimodal ConvNeXt model. The box boundaries represent the interquartile range (IQR). Ultrasound imaging contributed more strongly (mean 0.63 $\pm$ 0.09) than clinical features (mean 0.37 $\pm$ 0.09), while both provided complementary information for classification.}
    \label{fig:modality_ablation}
\end{figure}

We adopted a drop-based ablation approach to quantify the relative importance of each modality. For each sample, we computed the model’s predicted probability for the malignant class using all validation data across 5 folds. We repeated the process after zeroing out either the image or the clinical features. The magnitude of the change in probability is treated as the modality’s contribution. We normalized these changes to obtain relative contribution scores per sample. Figure \ref{fig:modality_ablation} illustrates that BUS contributes 63\% $\pm$ 0.09 in the class prediction and clinical data contributes to 37\% $\pm$ 0.09. The box plot boundaries represent the 25th and 75th percentiles.

\subsection{Performance comparison between unimodal and multimodal approaches}

We investigated the contribution of each modality by evaluating unimodal and multimodal models using our best-performing image encoder (ConvNeXt) and a multilayer perceptron (MLP) for clinical data. Table~\ref{tab:modality_comparison} summarizes the classification performance across different input configurations. The multimodal model performed best, with an accuracy of 0.9098, F1-score of 0.6720, and AUC of 0.9427, demonstrating the benefit of combining data sources. The image-only ConvNeXt also performed well (AUC 0.8919), while the clinical-only MLP was much weaker with a F1-score of 0.33, showing that clinical features alone have limited discriminative power. Overall, these findings confirm the value of multimodal fusion for improving diagnostic accuracy and robustness.




\begin{table}[!ht]
\caption{Performance comparison of different input modalities}
\label{tab:modality_comparison}
\centering
\begin{tabular}{@{}ccccc@{}}
\toprule
Modality    & Accuracy  & F1-Score  & AUC & 95\% CI (AUC) \\ \midrule
Only clinical data        & 0.8140 & 0.3300 & 0.7846 & [0.6561, 0.9132] \\
Only BUS image      & 0.8906 & 0.6138 & 0.8919 & [0.8259, 0.9578] \\
\textbf{Multimodal}     & \textbf{0.9098} & \textbf{0.6720} & \textbf{0.9427} & \textbf{[0.8842, 1.0000]} \\ \bottomrule
\end{tabular}
\end{table}

\begin{figure}[!ht]
    \centering
    \includegraphics[width=0.7\linewidth]{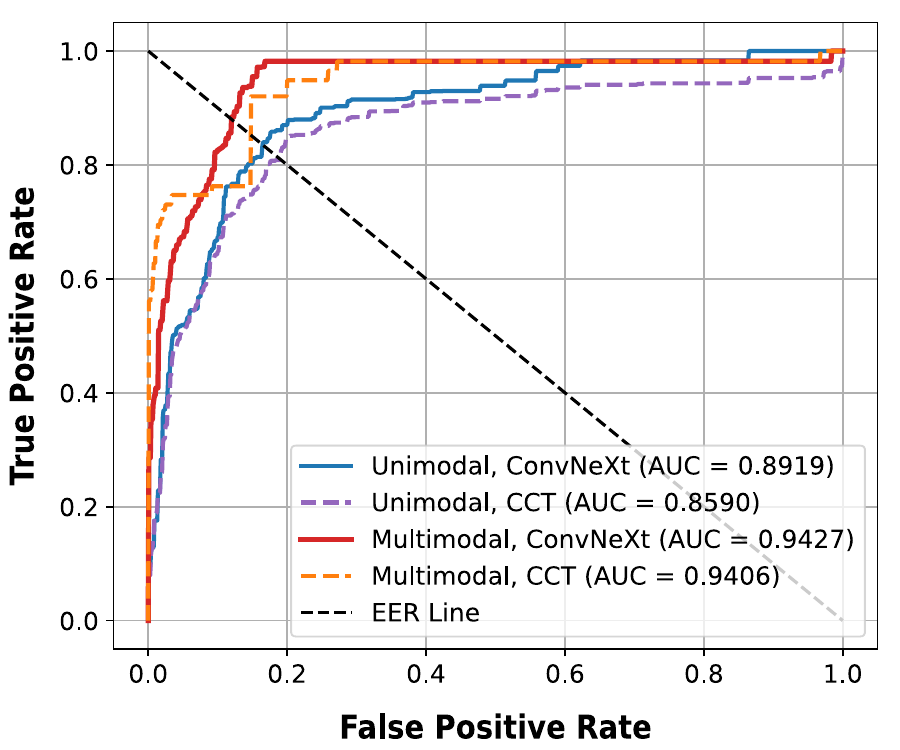}
    \caption{Receiver operating characteristic (ROC) curves of ConvNeXt and CCT in unimodal and multimodal settings. Multimodal models outperformed unimodal counterparts (ConvNeXt: 0.9427 vs. 0.8919; CCT: 0.9406 vs. 0.8590), confirming the benefit of integrating clinical data. The multimodal ConvNeXt achieved the highest AUC (0.9427) and EER (0.8799), followed by multimodal CCT (AUC 0.9406, EER 0.8519), with unimodal ConvNeXt (0.8348) and CCT (0.8098) performing lower.}
    \label{fig:roc_comparison}
\end{figure}

Figure~\ref{fig:roc_comparison} shows ROC curves of the two best-performing models, namely ConvNeXt and CCT, in both unimodal and multimodal configurations. The multimodal ConvNeXt model achieved the highest AUC of 0.9427 (95\% CI: [0.8842, 1.0000]), followed closely by multimodal CCT with an AUC of 0.9406 (95\% CI: [0.8633, 1.0000]), both of which surpassed their unimodal counterparts. At the EER point, multimodal ConvNeXt achieved the highest score (0.8799), followed by multimodal CCT (0.8519), both surpassing unimodal ConvNeXt (0.8348) and CCT (0.8098). This reiterates the value of multimodal fusion, especially in cases where image features alone may be insufficient for confident classification.

\section{Discussion}
While unimodal models based solely on ultrasound imaging achieved reasonably strong performance, the multimodal approach consistently outperformed them across all key metrics. Although the modality ablation study indicates a greater contribution from BUS images, the clinical data still plays a significant complementary role. This aligns with clinical practice, where both imaging and patient history are considered. 

Moreover, Score-CAM visualization provides a foundation for clinical interpretability. Misclassified cases tended to exhibit weaker or diffused attention maps, indicating that the image encoder struggled with ambiguous visual patterns, which further justifies the use of auxiliary modalities.

Despite promising results, this study has a few limitations. The phyllodes dataset is relatively small for malignant and borderline cases, which limits the statistical strength of the findings. Larger and more diverse multi-center datasets are needed before the clinical trial to further refine and test the AI models. Overall, this work shows that multimodal deep learning frameworks can improve phyllodes tumor diagnosis, even when working with small and imbalanced datasets which is a common challenge in rare diseases.

\section{Conclusion}
We proposed a multimodal deep learning framework that integrates ultrasound image features with structured clinical data for phyllodes tumor classification. The multimodal architecture consistently outperformed unimodal baselines across multiple image encoders, improving diagnostic accuracy and robustness while potentially reducing unnecessary surgical biopsies. A key challenge was the strong class imbalance, with relatively few borderline and malignant cases, which we mitigated using class-aware sampling and subject-stratified cross-validation.

Future work will aim to expand the dataset, incorporate radiologist annotations, and explore transformer-based encoders with attention mechanisms to enhance interpretability and performance. Prospective validation in real-world clinical settings will also be critical to assess the system’s generalizability and its utility in guiding biopsy decisions.

\bibliographystyle{IEEEtran}
\bibliography{bibtext/ref}{}

\end{document}